\documentclass[conference]{IEEEtran}
\IEEEoverridecommandlockouts
\usepackage{cite}
\usepackage{amsmath,amssymb,amsfonts}
\usepackage{graphicx}
\usepackage{textcomp}
\usepackage{xcolor}

\usepackage{algpseudocode}
\usepackage{algorithm}

\def\BibTeX{{\rm B\kern-.05em{\sc i\kern-.025em b}\kern-.08em
    T\kern-.1667em\lower.7ex\hbox{E}\kern-.125emX}}
\begin{document}

\title{AMMSM: Adaptive Motion Magnification and Sparse Mamba for Micro-Expression Recognition}

\author{
\IEEEauthorblockN{Xuxiong Liu$^{1,*}$, Tengteng Dong$^{1,*}$, Fei Wang$^{1,2}$,Weijie Feng$^{1}$, Xiao Sun$^{1,{\dagger}}$}
\IEEEauthorblockA{$^1$ School of Computer Science and Information Engineering, Hefei University of Technology, Hefei, China}
\IEEEauthorblockA{$^2$ Institute of Artificial Intelligence, Hefei Comprehensive National Science Center, Hefei, China}

     {\small \tt \{liuxuxiong, 2022111070\}@mail.hfut.edu.cn, jiafei127@gmail.com, \{wjfeng, sunx\}@hfut.edu.cn}
     \thanks{This work was supported by Special Project of the National Natural Science Foundation of China (62441614), Anhui Province Key R\&D Program (202304a05020068) and General Programmer of the National Natural Science Foundation of China (62376084).}
     \thanks{*Equal contribution. $^{\dagger}$Corresponding author.}
}

\maketitle

\begin{abstract}
Micro-expressions are typically regarded as unconscious manifestations of a person’s genuine emotions.
However, their short duration and subtle signals pose significant challenges for downstream recognition.
We propose a multi-task learning framework named the 
Adaptive Motion Magnification and Sparse Mamba (AMMSM)
to address this. 
This framework aims to enhance the accurate capture of micro-expressions through self-supervised subtle motion magnification, while the sparse spatial selection Mamba architecture combines sparse activation with the advanced Visual Mamba model to model key motion regions and their valuable representations more effectively.
Additionally, we employ evolutionary search to optimize the magnification factor and the sparsity ratios of spatial selection, followed by fine-tuning to improve performance further.
Extensive experiments on two standard datasets demonstrate that the proposed AMMSM achieves state-of-the-art (SOTA) accuracy and robustness.

\end{abstract}

\begin{IEEEkeywords}
Micro-expression recognition, Motion magnification, Sparse activation, Visual Mamba
\end{IEEEkeywords}
\section{Introduction}
\label{sec:intro}

Facial expressions are vital cues for understanding human emotions and intentions, serving as important nonverbal signals in both social or human-computer interaction. Facial expressions are generally classified into two categories: macro-expression and micro-expression~\cite{li2022deep}.
Macro-expressions are overt and sustained, but they can be consciously controlled or suppressed, which may undermine their reliability as authentic indicators of emotion. In contrast, micro-expressions are brief and subtle, and their typically involuntary nature makes them more reliable in reflecting a person's genuine emotions~\cite{wang2025exploiting}.



However, the detection and recognition of micro-expressions face significant challenges due to their transience and subtlety. Previous studies~\cite{ben2021video} have mentioned that manual recognition of micro-expressions is time-consuming and error-prone, and it is essential to develop MER systems with robust generalization capacities and high recognition accuracy.


In recent years, a series of influential studies have been presented~\cite{Wang_Li_Liu_Yan_Ou_Huang_Xu_Fu_2018, Khor_See_Phan_Lin_2018,nie2021geme, nguyen2023micron, fan2023selfme,gan2024laenet,zhou2022feature}, which can be broadly categorized into two approaches based on their input types: sequence-based methods~\cite{Wang_Li_Liu_Yan_Ou_Huang_Xu_Fu_2018, Khor_See_Phan_Lin_2018,nie2021geme} employ image sequences as input and motion-based methods~\cite{nguyen2023micron, fan2023selfme,gan2024laenet,zhou2022feature} utilize onset and apex frames to derive optical flow or extract motion-related features.
The primary challenge in MER stems from the subtlety of facial movements during micro-expressions, which provide insufficient features for reliable classification. Consequently, several studies have introduced motion magnification techniques~\cite{wang2024frequency,wang2024eulermormer}. 
However, these methods typically either use traditional image processing techniques for motion magnification or incorporate a pre-trained motion magnifier into the recognition pipeline.
These approaches encounter difficulties in determining the appropriate magnification factor and compromise the robust generalization capability of end-to-end models. Additionally, existing models based on the Transformer architecture are often over-parameterized, introducing considerable computational redundancy and overhead.

To address the aforementioned limitation, we propose a novel end-to-end framework for MER that integrates adaptive motion magnification technology. Furthermore, we introduce sparse activation into the Mamba model, resulting in Sparse Mamba, which enables spatial selection of key motion regions, thereby enhancing the model's recognition accuracy.
Experiments conducted on two standard datasets demonstrate that our model achieves SOTA performance. Furthermore, ablation studies were performed to validate the effectiveness of our approach.

Our contributions can be summarized as follows:
\begin{itemize}
  \item We propose a novel end-to-end MER model with adaptive motion magnification to address the challenge of detecting subtle facial movements. Unlike previous approaches, our model is designed to simultaneously minimize both magnification and classification loss, and it can adaptively select the appropriate magnification factor.
  \item We introduce the Sparse Mamba model, which automatically enables spatial selection of key facial motion regions, minimizing the impact of movements unrelated to MER.
  \item The proposed AMMSM framework is designed for multi-task learning and trained using an end-to-end deep network architecture. It achieves SOTA performance on two standard micro-expression benchmarks, including CASME II and SAMM.
\end{itemize}
\section{Related Work}
\subsection{Micro-Expression Recognition}
The objective of MER is to classify micro-expressions from a given image sequence. Due to their brief duration and subtle intensity, the features of these motions are very weak. 
Traditional methods achieve recognition by designing and utilizing handcrafted features, such as dual-cross patterns from three orthogonal planes (DCP-TOP)~\cite{Ben_Jia_Yan_Zhang_Meng_2018}.
In~\cite{patel2016selective}, one of the earliest deep learning-based methods for MER was proposed; this method utilizes multiple pre-trained networks to extract micro-expression features and subsequently applies evolutionary learning for feature selection. Wang et al.~\cite{Wang_Li_Liu_Yan_Ou_Huang_Xu_Fu_2018} proposed a transferring long-term convolutional neural network (TLCNN) to extract facial features from each frame of the video clip. Khor et al.~\cite{Khor_See_Phan_Lin_2018} encoded each frame into a feature vector using a CNN, which were then fed into a Long Short-Term Memory (LSTM) network to predict micro-expressions. However, methods that use entire image sequences as input generally exhibit lower accuracy compared to those that rely only on the onset and the apex frames, leading to the emergence of research based on the latter approach~\cite{chen2024prototype,liu2024micro,gan2024laenet,zhou2022feature}. With the increasing focus on Vision Transformer, researchers have progressively turned to using Transformer-based or hybrid architectures as backbones~\cite{nguyen2023micron,fan2023selfme}, resulting in enhanced performance.

\subsection{Motion Magnification}
One of the most significant challenges in MER is detecting subtle facial movements, which makes motion magnification techniques essential. Previous methods have utilized Eulerian motion magnification~\cite{wang2017effective} or Global Lagrangian motion magnification~\cite{le2018micro}, showing that enhancing motion intensity can lead to improved MER performance. With the introduction of learning-based motion magnification~\cite{wang2024eulermormer,wang2024frequency}, Lei et al.~\cite{lei2020novel} employed transfer learning to integrate this technique, which helped reduce noise introduced by motion magnification and improve recognition accuracy. Recently, a simple but powerful motion magnification method was proposed~\cite{pan2024self}, inspiring the design of our motion magnification module.

\subsection{Mamba framework}
Recently, State Space Models, particularly Mamba~\cite{gu2023mamba} and Mamba2~\cite{dao2024transformers,zhao2025temporal}, have attracted significant attention due to their efficiency in processing sequential information. With their linear computational complexity and excellent performance, researchers have begun applying these models to two-dimensional image data, leading to the development of several models, such as VMamba~\cite{liu2024vmamba} and VSSD~\cite{shi2024vssd}. In this work, we introduce the Visual Mamba to the MER task, aiming to improve both performance and efficiency.                                                                                                                                                                                                                                                                                                                                                                                                                                                                                                                                                                                                                                                                                                                                                                                                                                                                                                                                                                                                                                                                                                                                                                                                                                                                                                                                                                                                                                                                                                                                                                                                                                                                                                                                                                                                                                                                                                                                                                                                                                                                                                                                                                                                                                                                                                                                                                                                                                                                                                                                                                                                                                                                                                                                                                                                                                                                                                                                                                                                                                                                                                                                                                                                                                                                                                                                                                                                                                                                                                                                                                                                                                                                                                                                                                                                                                                                                                                                                                                                                                                                                                                                                                                                                                                                                                                                                                                                                                                                                                                                                                                                                                                                                                                                                                                                                                                                                                                                                                                                                                                                                                                                                                                                                                                                                                                                                                                                                                                                                                                                                                                                                                                                                                                                                                                                                                                                                                                                                                                                                                                                                                                                                                                                                                                                                                                                                                                                                                                                                                                                                                                                                                                                                                                                                                                                                                                                                                                                                                                                                                                                                                                                                                                                                                                                                                                                                                                                                                                                                                                                                                                                                                                                                                                                                                                                                                                                                                                                                                                                                                                                                                                                                                                                                                                                                                                                                                                                                                                                                                                                                                                                                                                                                                                                                                                                                                                                                                                                                                                                                                                                                                                                                                                                                                                                                                                                                                                                                                                                                                                                                                                                                                                                                                                                                                                                                                                                                                                                                                                                                                                                                                                                                                                                                                                                                                                                                                                                                                                                                                                                                                                                                                                                                                                                                                                                                                                                                                                                                                                                                                                                                                                                                                                                                                                                                                                                                                                                                                                                                                                                                                                                                                                                                                                                                                                                                                                                                                                                                                                                                                                                                                                                                                                                                                                                                                                                                                                                                                                                                                                                                                                                                                                                                                                                                                                                                                                                                                                                                                                                                                                                                                                                                                                                                                                                                                                                                                                                                                                                                                                                                                                                                                                                                                                                                                                                                                                                                                                                                                                                                                                                                                                                                                                                                                                                                                                                                                                                                                                                                                                                                                                                                                                                                                                                                                                                                                                                                                                                                                                                                                                                                                                                                                                                                                                                                                                                                                                                                                                                                                                                                                                                                                                                                                                                                                                                                                                                                                                                                                                                                                                                                                                                                                                                                                                                                                                                                                                                                                                                                                                                                                                                                                                                                                                                                                                                                                                                                                                                                                                                                                                                                                                                                                                                                                                                                                                                                                                                                                                                                                                                                                                                                                                                                                                                                                                                                                                                                                                                                                                                                                                                                                                                                                                                                                                                                                                                                                                                                                                                                                                                                                                                                                                                                                                                                                                                                                                                                                                                                                                                                                                                                                                                                                                                                                                                                                                                                                                                                                                                                                                                                                                                                                                                                                                                                                                                                                                                                                                                                                                                                                                                                                                                                                                                                                                                       
                                                                                                                                                                                                                                                                                                                                                                                                                                                           
\begin{figure}[t]
\centering
\includegraphics[width=0.45\textwidth]{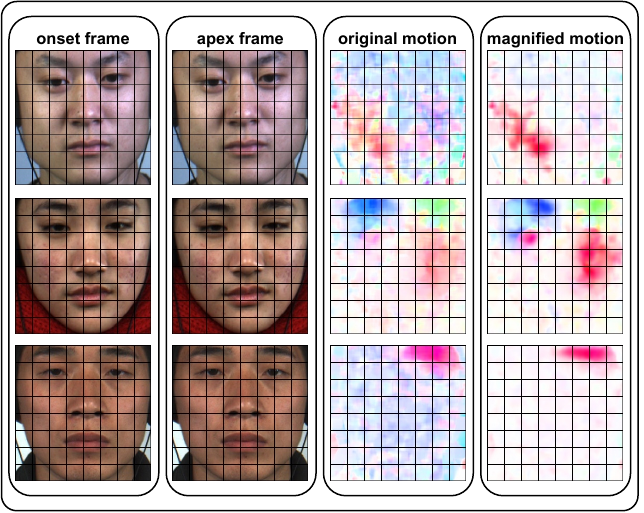}
\caption{\textbf{Visualization of magnified optical flow.} The colors of the motion maps represent the directions and intensity of the motion. The expressions from top to bottom are happiness, surprise, and disgust. The motion magnification module magnifies movement in regions such as the eyebrows, cheeks, and corners of the mouth while suppressing irrelevant motions for MER.}
\label{fig:flow}
\end{figure}

\begin{figure*}[t]
\centering
\includegraphics[width=0.98\textwidth]{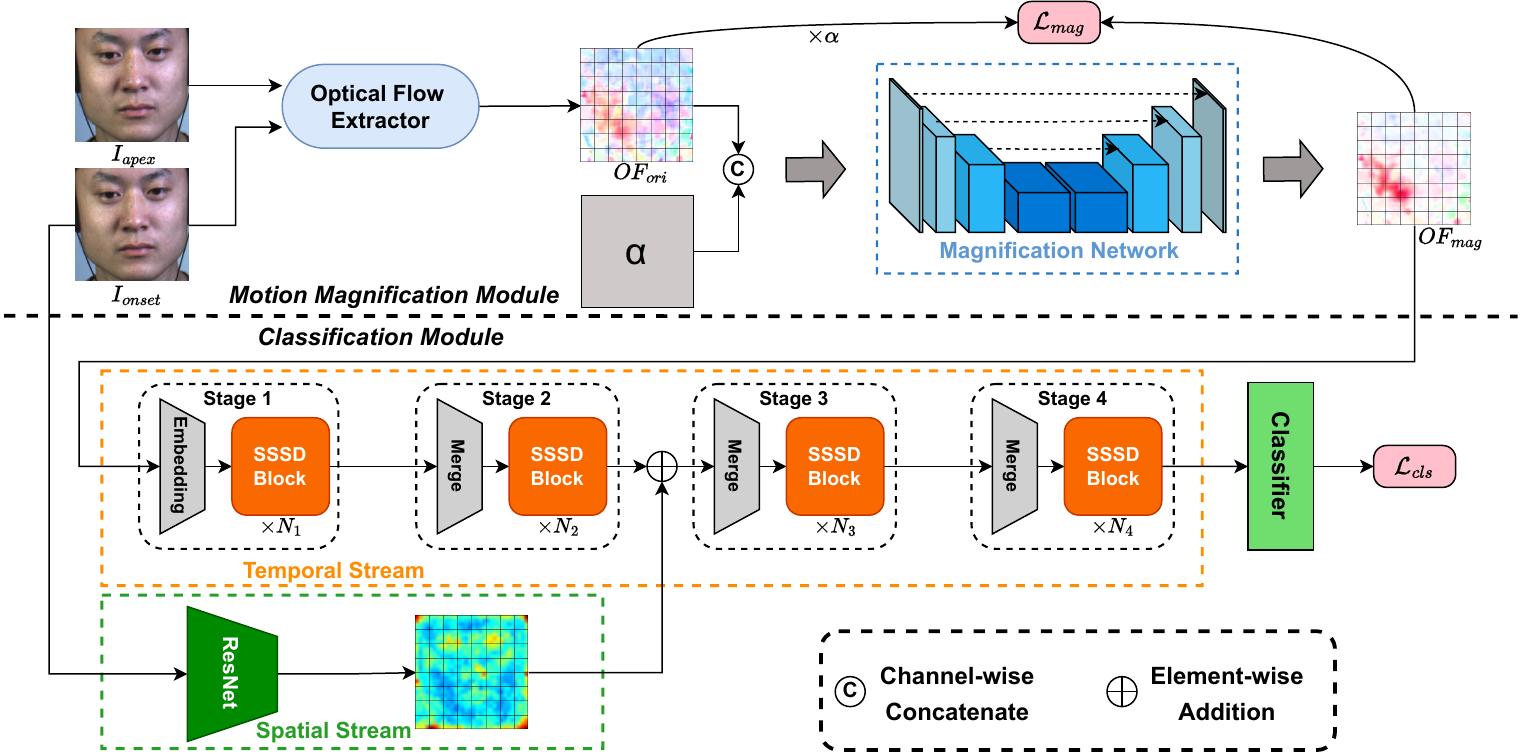}
\caption{\textbf{Overall architecture of the AMMSM Model.} AMMSM consists of a motion magnification module and a classification module. A standard UNet~\cite{du2025improving} is used as the motion magnifier within the motion magnification module. The classification module follows a two-stream network architecture. In the spatial stream, a ResNet18~\cite{he2016deep} extracts spatial features from $I_{onset}$. The temporal stream takes $OF_{mag}$ as input, with the backbone structured into four hierarchical stages, and the last layer of the last two blocks is replaced with MSA. The spatiotemporal feature fusion is performed at the end of stage 2.}
\label{fig:framework}
\end{figure*}

\section{Method}
In this work, we propose a novel end-to-end framework that leverages self-supervised learning to magnify facial motion while making predictions simultaneously, as shown in Figure~\ref{fig:framework}. Our model consists of a motion magnification module and a classification module. In the motion magnification module, the motion magnifier is trained on optical flow by integrating the magnification objective with the classification task, enabling it to focus exclusively on and magnify key regions relevant to MER. The classification module comprises a temporal stream and a spatial stream. The temporal stream primarily focuses on facial motion features, while the spatial stream extracts facial structural features. By integrating these two streams at a mid-level, more meaningful representations are learned, leading to improved performance in the MER task.

\subsection{Adaptive Motion Magnification}
Given an onset-apex image pair $I=\left\{I_{onset},I_{apex}\right\}\in\mathbb{R}^{H\times W\times 3\times2}$, the motion field is obtained by computing optical flow with an off-the-shelf optical flow estimator $\mathcal{F}_{OF}(\cdot)$. Our objective is to generate a magnified motion field, achieved by a standard UNet, denoted as $\mathcal{F}_{mag}$, which takes the original optical flow $OF_{ori}$ and a map of magnification factor $\alpha$~\cite{pan2024self} as input and outputs the magnified optical flow $OF_{mag}$. We employ a magnification loss that encourages the generated optical flow to be scaled to $\alpha$ times the input optical flow. The magnification loss $\mathcal{L}_{mag}$ is computed as follows:
\begin{align}
  \mathcal{L}_{mag} &= \left\|\alpha \cdot OF_{ori} - OF_{mag}\right\|_{1}.
\end{align}
And the total loss function is defined as:
\begin{align}
  \mathcal{L} = \mathcal{L}_{cls}+\frac{e_r-e}{e_r}\mathcal{L}_{mag},
\end{align}
where $e$ represents the current epoch and $e_r$ denotes the total number of epochs.
By utilizing this loss function, we eliminate the need for additional training data and processes. However, relying solely on this loss function may not be sufficient to generate a magnified motion field that is beneficial for MER. Therefore, we integrate it with the standard multi-class cross-entropy loss $\mathcal{L}_{cls}$ for end-to-end training. During the early stages of training, the model adapts to different magnification factors by randomly sampling within a specified range. In the later stages, $\alpha$ is optimized along with the sparsity ratios through evolutionary search, followed by fine-tuning with a stable configuration to improve performance.

To visually demonstrate the effect of the magnification module, we present the generated optical flow in Figure~\ref{fig:flow}. Compared to the original, the magnified optical flow more effectively highlights the key motion regions.

\subsection{Sparse State Space Duality}
\begin{figure*}[t]
\centering
\includegraphics[width=\textwidth]{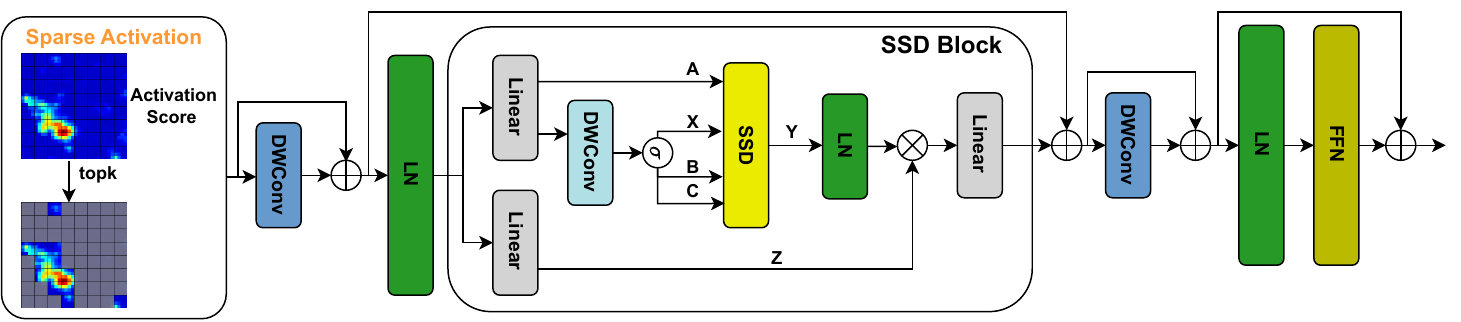}
\caption{\textbf{Architecture of SSSD block.} The SSSD block comprises a sparse activation module, an SSD block, and an FFN. The sparse activation module enables the SSD block to focus exclusively on the most critical regions of motion, thereby preventing the introduction of irrelevant information. Meanwhile, DWConv and FFN are used separately to enhance the model's ability to capture local information and promote cross-channel information exchange.}
\label{fig:block}
\end{figure*}

Understanding and recognizing facial movements is crucial for the MER task. To this end, we integrated the State Space Duality (SSD) model with sparse activation within the temporal stream of the classification module, introducing the Sparse State Space Duality (SSSD) block to enhance the model's ability to capture key motion regions. This integration leads to superior performance compared to existing models. The architecture of the SSSD block is shown in Figure~\ref{fig:block}.

\textbf{Block Design.} We adopt the SSD block design from~\cite{shi2024vssd}, which replaces the causal 1D convolution with Depth-Wise Convolution (DWConv) in line with previous Visual Mamba works, and incorporates a DWConv before both the SSD block and the Feed-Forward Network (FFN) to enhance the model's capability for local feature perception. Furthermore, we introduce two modifications. First, we implement a sparse activation module that enables the spatial selection of regions with the highest activation scores, thereby preventing the influence of less important motion features on MER. Second, based on the findings of Mamba2~\cite{dao2024transformers}, which show that the SSD block and Multi-head Self-Attention (MSA) block are complementary, and integrating these two blocks can yield better performance, we replace the final SSD block in the last two stages with an MSA block to improve performance.

\textbf{Detail.} At the beginning of each stage, we calculate the importance score of each pixel as the basis of selection. To preserve the spatial integrity of the selected regions, we divide the feature map into $4\times4$ windows. To illustrate, we consider the first stage where the input $X\in\mathbb{R}^{H\times W\times C}$, the activation maps are obtained in the shape of windows as:
\begin{align}
  X_{window}=\left\{X_{m,n}\in\mathbb{R}^{4\times4\times C}\right\}_{m=0,n=0}^{m=\frac{H}{4}-1,n=\frac{W}{4}-1}.
\end{align}
Next, we evaluate the importance of each window $\Phi_{m,n}$ by computing its L2 norm:
\begin{align}
  \Phi_{m,n} = \left \|X_{m,n}  \right \| _2.
\end{align}
After determining the sparsity ratio $\mathcal{S}^i_j$ of the layer, where $i$ denotes the stage number and $j$ denotes the layer number within that stage, we derive a mask to identify the regions that will be selected:
\begin{align}
  M_{m,n} = 
\begin{cases} 
\mathbf{1}, & \text{if } X_{m,n} \in topk(\Phi,\mathcal{S}^i_j) \\
\mathbf{0}, & \text{otherwise} 
\end{cases},
\end{align}
where $topk(\Phi,\mathcal{S}^i_j)$ denotes the selection of the top $(1-\mathcal{S}^i_j)$ proportion of the largest values from $\Phi$.

Before entering the SSD block, less important features are masked by multiplying the input $X$ with the mask $M$. At the end of this stage, the features from the masked regions are directly copied back into the feature map to prevent information loss.

We adapt the setup from~\cite{chen2023sparsevit}, where the importance scores are calculated only at the beginning of each stage to minimize computational overhead, while the mask is recalculated whenever the sparsity ratio changes. Our experimental results indicate that maintaining a consistent sparsity ratio across every two blocks leads to optimal performance.

After aggregation by a DWConv, we obtain the input $u\in\mathbb{R}^{H\times W\times C}$ for the SSD module. The main function of the SSD module can be expressed as:
\begin{align}
  Y&=C(B^T(X\cdot \frac{1}{A})) ,
\end{align}
where $X,A,B,C$ are both generated by $u$ and SSD block's parameters.

Finally, the output of the SSD block is fed into a DWConv and an FFN, which refine the feature representation and produce the final result.

\subsection{Adaptive Training and Optimum Configuration Search}
To equip the model with the ability to predict under various sparsity ratios and magnification factors, we employ an adaptive training approach during the initial phase: randomly sampling these parameters within a specified range and training the model accordingly. In our experiments, this phase lasts for 70 epochs.
We then apply an evolutionary search approach to identify the optimal configuration for these parameters, using the classification loss $\mathcal{L}_{cls}$ as the fitness criterion.
Finally, the model is fine-tuned under this configuration for 30 epochs to maximize performance.

    
\section{Experiment}
\subsection{Dataset}
\textbf{CASME II}~\cite{yan2014casme}. CASME II has 35 participants and 247 micro-expression sequences. Videos are at 200 frames per second (FPS) and resolutions of $640\times480$ with $280\times340$ for the face. In our experiment, we applied two classification methods: a three-class and a five-class approach. The three-class method categorizes micro-expressions into positive, negative, and surprise, while the five-class method includes happiness, depression, disgust, surprise, and others.

\textbf{SAMM}~\cite{davison2016samm}. SAMM has a 200 FPS frame rate and a facial resolution of 400 × 400. It consists of 159 samples from 32 participants and 13 ethnicities. Each sample includes an emotion label, an apex frame, and action unit annotations. Similar to the three-class classification in CASME II, micro-expressions are categorized into positive, negative, and surprise.

\subsection{Performance Metrics}
We adopted the standard metrics and protocols proposed by the MER2019 challenge~\cite{see2019megc}. All experiments were conducted using a leave-one-subject-out (LOSO) cross-validation approach, where one subject was selected for testing at a time, and the remaining samples were used for training. Performance was measured using the unweighted F1-score (UF1) and unweighted average recall (UAR). The UF1 is calculated as:
\begin{align}
  UF1=\frac{1}{C}\sum_{i=1}^{C}\frac{2\cdot TP_i}{2\cdot TP_i + FP_{i}+FN_i},
\end{align}
and UAR is defined as:
\begin{align}
UAR=\frac{1}{C}\sum_{i=1}^{C}\frac{TP_i}{n_i},
\end{align}
where $C$ represents the number of micro-expression classes, $n_i$ is the number of samples of $i$-th class, and $TP_i$, $FP_i$, and $FN_i$ are true positives, false positives, and false negatives for the $i$-th class, respectively.

\subsection{Implementation Details}
AdamW optimizer~\cite{du2025improving} is applied with an initial learning rate of $3\times10^{-5}$ and a batch size of 16. The learning rate is managed by a cosine decay scheduler, combined with a weight decay rate of 0.05. MESA~\cite{mesa} and Mixup~\cite{mixup} are employed to further enhance the model's generalization. The sparsity ratios are selected from \{10\%, 20\%, ..., 80\%\}, and the magnification factor ranges from 1 to 4. The classification model consists of four stages, with layer counts of [2, 4, 8, 4] and channel sizes of [64, 128, 256, 512]. An NVIDIA RTX A6000 GPU is utilized for all experiments.

\subsection{Comparison with the SOTA Methods}
We compared our AMMSM method with mainstream deep learning approaches on two standard datasets. Table~\ref{tab:casme} details results for CASME II, where our method demonstrates improvements over existing approaches. For three categories, our method achieves a UF1 score of 93.76\% and UAR of 93.45\%, represent 2.75\% and 0.55\% increases over previous leading methods, respectively. For the five-class classification, it achieves a UF1 score of 85.23\% and UAR of 84.48\%. While the UF1 score is comparable to the previous SOTA method, the UAR shows an improvement of 1\%.

The results on SAMM are shown in Table~\ref{tab:samm}. Our method achieves a UF1 score of 76.40\% and a UAR of 76.22\%. While the UF1 score is 1.14\% lower than that of previous methods, there is a 4.67\% improvement in UAR.
\begin{table}[t]
  \begin{center}
    \caption{MER on CASME II dataset} \label{tab:casme}
    \begin{tabular}{|c|c|c|c|}
      \hline
      Method & Classes & UF1(\%) & UAR(\%)
      \\
      \hline
      CapsuleNet(2019)~\cite{capsuleNet}     & 3     & 70.68  & 70.18     \\
      EMRNet(2019)~\cite{liu2019EMR}    & 3     & 82.93  & 82.09     \\
      FR(2022)~\cite{zhou2022feature} & 3     & 89.15  & 88.73     \\
      $\mu$-BERT(2023)~\cite{nguyen2023micron}     & 3     & 90.34  & 89.14     \\
      SelfME(2023)~\cite{fan2023selfme} & 3     & 90.78  & 92.90     \\
      LAENet(2024)~\cite{gan2024laenet}     & 3     & 91.01  & 91.19     \\
      \textbf{AMMSM(ours)}       & 3     & \textbf{93.76}  & \textbf{93.45}     \\
      \hline
      SMA-STN(2020)~\cite{liu2020sma}     & 5     & 82.59  & 79.46     \\
      GEME(2021)~\cite{nie2021geme}     & 5     & 73.54  & 75.20     \\
      $\mu$-BERT(2023)~\cite{nguyen2023micron}     & 5     & \textbf{85.53}  & 83.48     \\
      \textbf{AMMSM(ours)}       & 5     & 85.23   &\textbf{84.48}     \\
      \hline
    \end{tabular}
  \end{center}
\end{table}

\begin{table}[t]
  \begin{center}
    \caption{MER on SAMM dataset} \label{tab:samm}
    \begin{tabular}{|c|c|c|c|}
      \hline
      Method & Classes & UF1(\%) & UAR(\%)
      \\
      \hline
      CapsuleNet(2019)~\cite{capsuleNet}     & 3     & 62.09  & 59.89     \\
      EMRNet(2019)~\cite{liu2019EMR}     & 3     & \textbf{77.54}  & 71.52     \\
      FR(2022)~\cite{zhou2022feature}    & 3     & 73.72  & 71.55     \\
      LAENet(2024)~\cite{gan2024laenet}     & 3     & 68.14  & 66.20     \\
      \textbf{AMMSM(ours)}       & 3     & 76.40  & \textbf{76.22}    \\
      \hline
    \end{tabular}
  \end{center}
\end{table}

\begin{table}[t]
  \begin{center}
    \caption{Ablation study of key component. } \label{tab:ablation}
    \begin{tabular}{|c|c|c|c|c|l|}
      \hline
      AMM & SA & Backbone & UF1(\%) & UAR(\%) & Latency(ms)
      \\
      \hline
      -              & -              & Transformer  & 77.42   &74.66 & 5.79\\
      \checkmark     & -              & Transformer  & 82.25   &80.94 & - \\ 
      -              & \checkmark     & Transformer  & 77.86   &74.87 & - \\ 
      \checkmark     & \checkmark     & Transformer  & 84.94   &83.42 & 9.56\\ 
      \hline
      -              & -              & Mamba        & 73.84   &70.85 & 3.21\\ 
      \checkmark     & -              & Mamba        & 82.94   &80.94 &  - \\  
      -              & \checkmark     & Mamba        & 74.97   &72.84 &  - \\  
      \hline
      \checkmark     & \checkmark     & Mamba        & \textbf{85.23} &\textbf{84.48} & \textbf{6.92}~\textcolor{red}{\scriptsize ↑27.6\%}\\
      \hline
    \end{tabular}
  \end{center}
\end{table}

\subsection{Ablation Study}
Experiments were conducted on CASME II to evaluate the contribution of the Adaptive Motion Magnification (AMM) module, the Sparse Activation (SA) module, and the Mamba backbone. As shown in Table~\ref{tab:ablation}, both the Transformer model and the Mamba model exhibit significant improvements in UAR and UF1 scores with the implementation of the AMM module.
The SA module also contributes to performance enhancements for both backbones.
In terms of computational resources, the Mamba model incurs only 72.4\% of the latency required by the Transformer model during batch training while maintaining comparable performance.

\begin{figure}[t]
\centering
\includegraphics[width=0.45\textwidth]{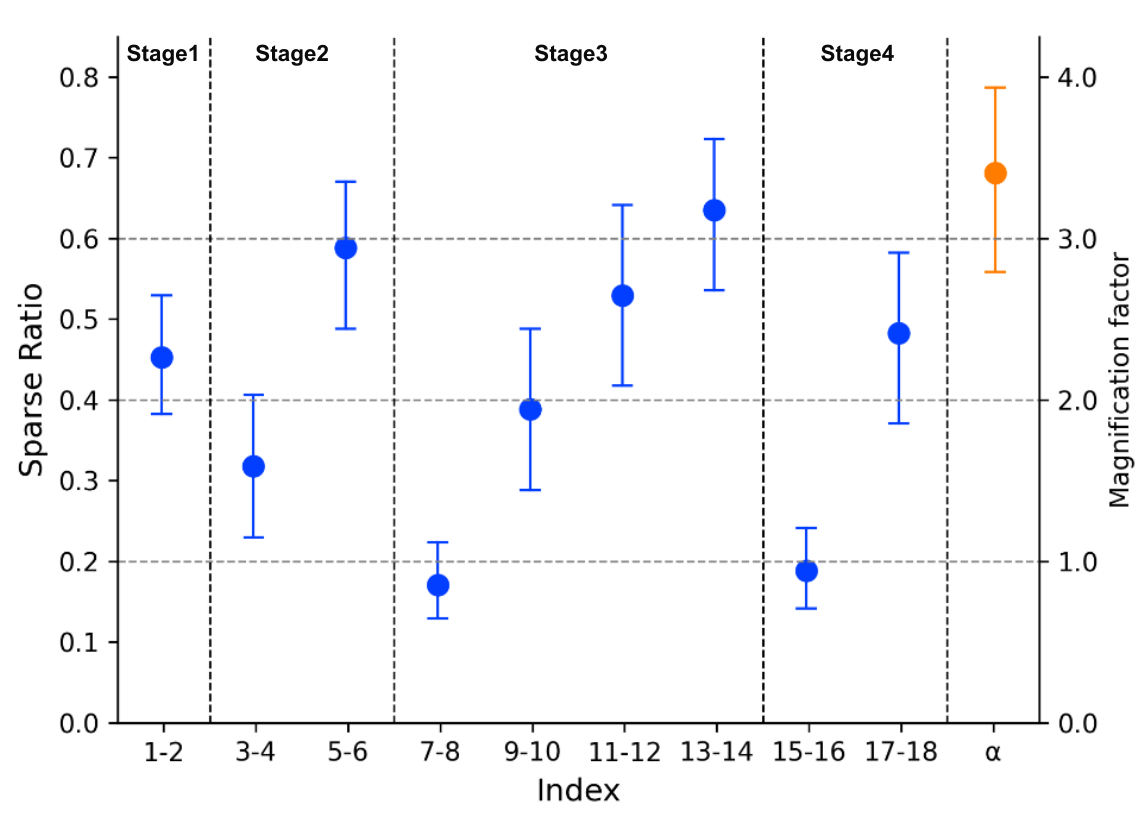}
\caption{\textbf{Distribution of sparsity ratios and magnification factor.} The blue dots represent the distribution of sparsity ratios, while the orange dot represents the distribution of the magnification factor.}
\label{fig:ratio}
\end{figure}

\subsection{The role of Sparse and Magnification}
The performance improvements of sparse activation and magnification have been validated in ablation experiments. Additionally, we analyzed the distribution of the optimized sparsity ratios and magnification factor obtained through evolutionary search, as shown in Figure~\ref{fig:ratio}. In the early stages of the network, the model tends to select a relatively aggressive sparsity ratio, while in the later stages, it gradually shifts from a lower to a higher sparsity ratio. We hypothesize that the high resolution of the feature maps in early stages results in many windows being irrelevant to MER. Therefore, excluding these irrelevant windows can benefit the model. In the later stages, after extensive feature aggregation, most windows are likely related to MER, leading the model to become more cautious and gradually increase the sparsity ratio. 

Regarding the magnification factor, the model typically selects a higher value, suggesting that motion magnification is beneficial for MER. However, due to the limited sample size of the dataset, generalizing to higher magnification levels through adaptive learning is challenging. We plan to address this limitation in future work with a larger dataset.
\section{conclusion}
In this study, we introduce the AMMSM framework for the MER task. AMMSM consists of a motion magnification module that enhances subtle facial motion and a Sparse Mamba model that selectively focuses on key regions of the magnified motion. An evolutionary search is then employed to identify the optimal configuration and fine-tune the model, ultimately achieving SOTA performance on the CASME II and SAMM datasets. The effectiveness of each component is further validated through ablation studies. In future work, we plan to explore more advanced generative models on larger datasets to further improve recognition accuracy and generalization.

\bibliographystyle{IEEEbib}
\bibliography{main}

\end{document}